\documentclass{article}

\PassOptionsToPackage{numbers,sort&compress}{natbib}
\usepackage[preprint]{ail_at_hku}
\usepackage[utf8]{inputenc}
\usepackage[T1]{fontenc}
\usepackage{hyperref}
\usepackage{url}
\usepackage{booktabs}
\usepackage{amsmath}
\usepackage{amsfonts}
\usepackage{nicefrac}
\usepackage{microtype}
\usepackage{xcolor}
\usepackage{multirow}
\usepackage{array} 
\usepackage{float}
\usepackage{graphicx}

\title{AtomEgo: Exploring Ego–Robot Integration for Embodied Foundation Model Pretraining}

\author{%
\begin{minipage}{\linewidth}
  \centering
  \normalsize\sffamily\bfseries
  \mbox{Di Wu\textsuperscript{1,4}}\quad
  \mbox{Dongchen Zheng\textsuperscript{1,3}}\quad
  \mbox{Junhe Sheng\textsuperscript{1}}\quad
  \mbox{Zhongxing Wei\textsuperscript{1}}\quad
  \mbox{Songxin Zhang\textsuperscript{2}}\\[1pt]
  \mbox{Zejian Xie\textsuperscript{2}} \quad
  \mbox{Xiaoquan Sun\textsuperscript{5}}\quad
  \mbox{Junyang Zheng\textsuperscript{1}}\\[1pt]
  \mbox{Zhuoyang Song\textsuperscript{2}}\quad
  \mbox{Jiaxing Zhang\textsuperscript{2}}\quad
  \mbox{Jiayu Chen\textsuperscript{1,3}}\\[8pt]
  \normalfont\normalsize\rmfamily
  \mbox{\textsuperscript{1}INFIFORCE}\quad
  \mbox{\textsuperscript{2}Lionrock Artificial Intelligence Laboratory}\\[3pt]
  \mbox{\textsuperscript{3}HKU}\quad
  \mbox{\textsuperscript{4}Tongji University}\quad
  \mbox{\textsuperscript{5}Huazhong University of Science and Technology}
\end{minipage}
}

\begin{document}

\maketitle

\begin{abstract}{\url{https://github.com/Agentic-Intelligence-Lab/Atom-0}}
                  {}
                  {Jiayu Chen (jiayuc@hku.hk)}
                  {Sep, 2026}

Embodied foundation models are constrained by the limited scale and diversity of robot demonstrations, motivating the use of large-scale egocentric human interaction data.
However, how to effectively incorporate such data into embodied-model pre-training remains unclear because of substantial embodiment and action-space gaps between humans and robots.
We present \textbf{AtomEgo}, a systematic study of ego--robot co-training supported by a curated corpus of approximately 2,659 hours and a scalable data processing  pipeline.
Across vision--language--action and world--action model architectures, we investigate three representative paradigms: joint co-training with domain-specific action heads, progressive ego-to-robot transfer through embodiment alignment, and joint video--action modeling.
We evaluate these paradigms through multi-task real-robot experiments and language-conditioned cross-embodiment representation analysis.
Our results reveal a simple principle: \emph{Data Scale $\times$ Alignment Quality $\rightarrow$ Capability Gain}; egocentric data can improve generalization, but their value depends on how effectively they are aligned and utilized.
This principle can provide practical guidance for scalable ego--robot pre-training.
\end{abstract}

\section{Introduction}

Recent vision--language--action (VLA) models combine pretrained vision--language representations with action generation, enabling instruction-conditioned manipulation and transfer across tasks and embodiments~\cite{kim2024openvla,pi07}. 
World--action models (WAMs) ~\cite{worldmodel,fastwam} further learn predictive physical dynamics from visual observations, offering a complementary route toward general-purpose embodied intelligence. 
These developments suggest that embodied foundation models may benefit from the same scaling principles that have driven progress in language and multimodal modeling~\cite{scaling}.  
However, scaling these models requires large and diverse datasets, while robot demonstrations are costly to collect, constrained by specific hardware, and limited in task, environment, object, and interaction diversity~\cite{datapyramid,OXE}.
Consequently, data has become a central bottleneck to scaling embodied foundation models.

Egocentric human interaction data provide a compelling source of scalable supervision. 
First-person videos capture rich object interactions in diverse, naturally occurring environments and can be collected with substantially less hardware and operational overhead than robot trajectories~\cite{nair2022r3m}. 
Yet their scale does not translate automatically into useful policy supervision. 
Most egocentric videos lack executable action labels, while inferred hand motions differ from robot commands in morphology, kinematics, coordinate frames, and visual appearance. 
Existing studies have demonstrated several promising mechanisms for bridging this gap, including action retargeting, embodiment alignment, and latent-action or world-model objectives~\cite{egomimic2024,egoscale2026,ego2robot2026,joyaira2026}.
Nevertheless, these methods have largely been studied in isolation, and many operate in post-training rather than large-scale pre-training. 
This naturally raises the question:

\begingroup
\colorlet{abscolback}{black!5}
\begin{abstractbox}
\centering
\begin{minipage}{0.92\linewidth}
\centering\textbf{How can egocentric data be incorporated simply and effectively into ego--robot co-training during pre-training to improve model performance?}
\end{minipage}
\end{abstractbox}
\endgroup

To answer this question, we present \textbf{AtomEgo}, a systematic study of ego-robot co-training for embodied foundation models. 
Rather than committing to a single mechanism for exploiting human data, we study both VLA and WAM architectures and investigate three complementary pre-training paradigms: joint co-training with domain-specific action heads (Section~\ref{sec:direct-utilization}), progressive ego-to-robot transfer through embodiment alignment (Section~\ref{sec:explicit-alignment}), and joint video--action modeling for cross-embodiment transfer (Section~\ref{sec:world-modeling}). 
Together, these paradigms test whether simply isolating domain-specific supervision is sufficient to bridge the human--robot gap, or whether effective transfer requires a stronger mechanism through explicit embodiment alignment or a unified visual-prediction objective.

To support this study, we curate an ego-robot corpus at the scale of 3,033 hours and develop a scalable data processing pipeline. 
We then conduct systematic pre-training experiments with matched data and computational settings, integrating each ego-data paradigm into a unified embodied foundation model. 
Finally, we build an evaluation pipeline that measures policy performance and generalization across tasks and conditions, enabling objective comparison among the three paradigms and yielding practical guidance for future ego-robot pre-training.
This guidance reveals a key principle: scaling egocentric data alone is insufficient; high-quality alignment is equally essential for converting human experience into robot capability.

Our main contributions are summarized as follows:
\begin{itemize}
    \item We construct a curated 3,033-hour ego--robot pre-training corpus with a scalable processing and quality-control pipeline for heterogeneous interaction data.
    \item We formulate and implement three paradigms for incorporating egocentric data into embodied foundation-model pre-training, spanning domain isolation, explicit embodiment alignment, and joint video--action modeling. 
    \item We conduct controlled pre-training and comprehensive evaluation under a unified data and experimental framework, enabling a systematic comparison of alternative ego-robot co-training strategies.
    \item We derive empirical takeaways for aligning and utilizing egocentric data and reveal a simple principle: \emph{Data Scale $\times$ Alignment Quality $\rightarrow$ Capability Gain}, linking capability gains to both scale and alignment.
\end{itemize}

\section{Related Work}
\subsection{Embodied Foundation Models}

Vision--language--action (VLA) models extend pretrained vision--language models with action generation, enabling instruction-conditioned policies trained on large, heterogeneous robot corpora~\citep{rt2,kim2024openvla,octo2024,pi_0,pi05}.
Recent work advances cross-embodiment transfer, long-horizon execution, temporal memory, and tactile grounding~\citep{tactilevla2025,lingbotvla2,beingh07,rmbench,pi07,atomvla,llava-vla}.
In parallel, world--action models (WAMs) jointly predict actions and future visual states \citep{worldmodel}, using video dynamics to complement VLA semantic priors~\citep{dreamzero2026,cosmos3,motus,fastwam,himem-wam}.
However, embodied foundation models remain limited by scarce, costly robot demonstrations~\citep{OXE,datapyramid,zhong2025survey}, and naively mixing incompatible embodiments and action spaces can cause negative transfer~\citep{qwenrobotmanip2026,joyaira2026}.
We therefore study how to align large-scale egocentric experience with robot data for effective embodied foundation-model pre-training.

\subsection{Leveraging Egocentric Data for Embodied Pre-Training}
Egocentric data are video recordings collected from a human wearer's viewpoint that capture first-person human interactions at scale~\citep{nair2022r3m,MVP}, but their human-centric motions are not directly executable by robots~\citep{egomimic2024,egoverse2026}.
Prior work bridges human observations and motions to robot representations and control, demonstrating the value of egocentric data across VLA and WAM formulations~\citep{lapa,egomimic2024,egovla2025,beingh0,egoscale2026,egopi2026,beingh07}.
However, how to jointly use ego and robot data during VLA/WAM pre-training stage remains a challenge. 
For example, Ego2Robot~\citep{ego2robot2026} converts human videos into robot-format visual--action trajectories, whereas JoyAI-RA 0.5~\citep{joyaira2026} aligns heterogeneous data through implicit latent actions and explicit canonical actions.
However, existing studies primarily validate individual design choices, rather than systematically comparing different ego--robot pre-training strategies under controlled settings \citep{joyaira2026,ego2robot2026,beingh0}. Our work systematically examines multiple co-training paradigms within a unified framework.


\section{Data Preprocessing and Curation}

This section describes our data recipe, data processing, and the quality-control procedure.
Starting from approximately 3,033 hours of robot and egocentric data, our quality filtering pipeline retains about 2,659 hours of trajectories. 

\subsection{Data Recipe}

Our final expanded co-training recipe contains 334,054 training episodes, corresponding to approximately 2,659 hours. We organize the corpus into three groups according to their roles in ego-robot co-training; Table~\ref{tab:data-recipe} reports the contribution of each source.

\textbf{Robot Data.} We combine open-source robot demonstrations from AgiBot World Beta~\citep{agibotworld2025}, DROID~\citep{droid2024}, RoboCOIN~\citep{robocoin2025}, and RoboMIND~\citep{robomind2024}. We also collect in-house demonstrations using the Piper bimanual robot. Together, these datasets span single-arm, bimanual, mobile, and humanoid robots, providing broad cross-embodiment action supervision over approximately 1,560 hours.

\textbf{Ego Data.} We use EgoVerse~\citep{egoverse2026} as the source of approximately 1,079.5 hours of large-scale egocentric interaction data. Its first-person human and robot trajectories provide diverse manipulation experience beyond the environments and tasks covered by the robot datasets.

\textbf{Alignment Data.} Beyond the in-house robot demonstrations, we collect egocentric human demonstrations using wearable devices, and record robot executions of the same task set. We refer to these human and robot trajectories as alignment data: although they differ in embodiment and action space, they share task semantics and interaction objectives, providing explicit supervision for bridging the human--robot embodiment gap. This collection contains 1,296 episodes, corresponding to approximately 20 hours.

\begin{figure*}[t]
\centering
\includegraphics[width=\textwidth]{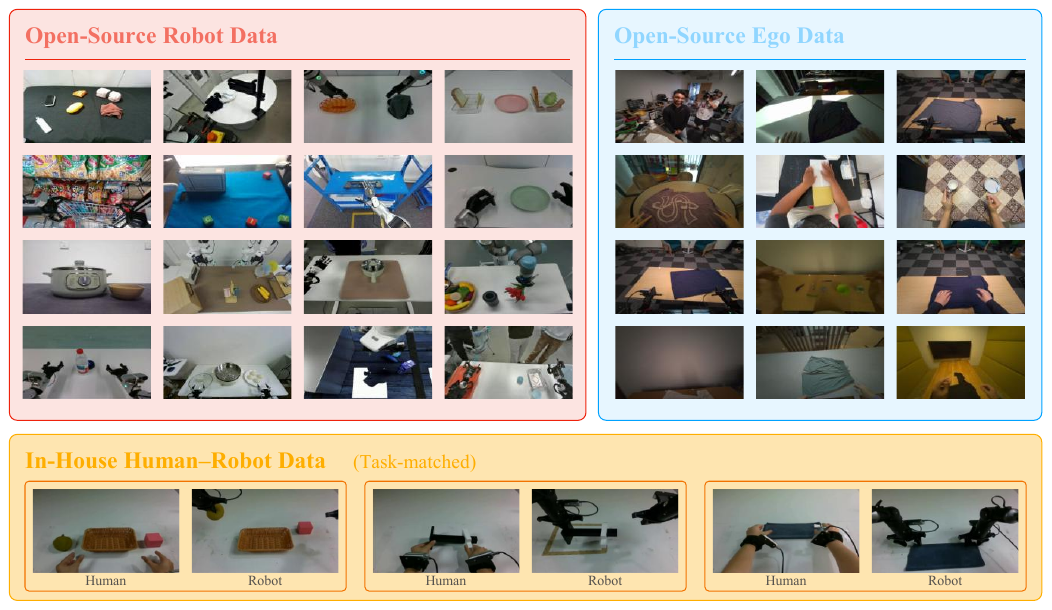}
\caption{Qualitative overview of the three data pools used in AtomEgo pre-training: open-source robot data, open-source egocentric data, and in-house task-matched human--robot alignment data.}
\label{fig:pretraining-data-overview}
\end{figure*}

\begin{table}[t]
\centering
\caption{Composition of the expanded co-training recipe. Proportions are calculated by training episode count.}
\label{tab:data-recipe}
\small
\begin{tabular}{llrrr}
\toprule
Group & Source & Episodes & Hours & Proportion \\
\midrule
\multirow{5}{*}{Robot data}
 & Piper (in-house) & 5,829 & 25.5 & 1.745\% \\
 & AgiBot World Beta~\citep{agibotworld2025} & 21,837 & 314.0 & 6.537\% \\
 & DROID~\citep{droid2024} & 64,124 & 343.0 & 19.196\% \\
 & RoboCOIN~\citep{robocoin2025} & 93,352 & 655.7 & 27.945\% \\
 & RoboMIND~\citep{robomind2024} & 83,152 & 221.7 & 24.892\% \\
\midrule
Ego data & EgoVerse~\citep{egoverse2026} & 64,464 & 1,079.5 & 19.297\% \\
\midrule
Alignment data & In-house aligned ego--robot data & 1,296 & $\sim$20.0 & 0.388\% \\
\midrule
Total & & 334,054 & $\sim$2,659.5 & 100\% \\
\bottomrule
\end{tabular}
\end{table}

\begin{table}[t]
\centering
\caption{Semantic layout of the unified 80-dimensional state--action space. Indices are zero-based and end-exclusive.}
\label{tab:action-space}
\small
\begin{tabular}{llll}
\toprule
Range & Width & Physical meaning & Action semantics \\
\midrule
$0{:}7$ & 7 & Left arm joints & Relative \\
$7{:}13$ & 6 & Left end-effector position and Euler rotation & Absolute \\
$13{:}16$ & 3 & Reserved & Masked \\
$16$ & 1 & Left gripper & Absolute \\
$17{:}29$ & 12 & Left hand joints & Absolute \\
$29{:}36$ & 7 & Right arm joints & Relative \\
$36{:}42$ & 6 & Right end-effector position and Euler rotation & Absolute \\
$42{:}45$ & 3 & Reserved & Masked \\
$45$ & 1 & Right gripper & Absolute \\
$46{:}58$ & 12 & Right hand joints & Absolute \\
$58{:}70$ & 12 & Left and right leg joints & Absolute \\
$70{:}74$ & 4 & Head and waist joints & Absolute \\
$74{:}75$ & 1 & Other body joint & Absolute \\
$75{:}80$ & 5 & Reserved & Masked \\
\bottomrule
\end{tabular}
\end{table}

\subsection{Data quality control}

Inspired by Qwen-Manip~\citep{qwenrobotmanip2026}, we apply three complementary state--action checks to identify corrupted or inconsistent episodes before training. Applying this quality-control procedure to the original 3,033 hours of data yields the final 2,659-hour training recipe.

\textbf{Sudden-change detection (S1).} We smooth each state and action dimension and detect abrupt changes using the residual, acceleration, and jerk relative to robust split-level thresholds. 
A frame is flagged only when a large residual coincides with abnormal acceleration or jerk, reducing false positives from normal rapid motion.

\textbf{State--action trend alignment (S2).} For physically comparable state--action dimensions, we measure temporal lag through cross-correlation and evaluate whether their motion directions agree. 
This check identifies mismatched trajectories, timing offsets, dropped actions, and cases in which the observed state incorrectly leads the command.

\textbf{Extreme-value filtering (S3).} We estimate per-dimension 1st and 99th percentiles over each dataset and flag values outside an expanded quantile interval, as well as any NaN or infinite values. 
This removes severe long-tail outliers that could distort normalization and destabilize training.

\subsection{Data processing}

\textbf{Format unification.} We convert each filtered source dataset into a validated episode-level RLDS representation. 
Distinct embodiment, state-action space, and camera configurations are retained as separate builders, producing 45 builders in total. At loading time, all builders are mapped to a common model interface, with unavailable camera views replaced by masked placeholders.

\textbf{Unified state--action space.} The shared 80-dimensional space is organized by physical semantics rather than by simply padding each native vector. 
The principal slots are listed in Table~\ref{tab:action-space}. 
Each dataset defines a mapping from its native state and action fields to these slots. 
Dimensions not used by a sample are set to zero and excluded from training through the action mask. 
Single-arm embodiments are consistently mapped to the right arm and gripper slots.

For egocentric demonstrations, we use a 12-dimensional bimanual action representation in which each hand is encoded as an absolute end-effector pose comprising 3D position and Euler angles.

For arm slots configured as relative joint actions, the target at horizon step $h$ is computed against the current observation state,
\begin{equation}
\Delta a_{t,h,j}=a_{t+h,j}-s_{t,j}.
\end{equation}
Other valid slots, including gripper commands, dexterous-hand joints, body joints, and end-effector poses, retain their absolute semantics. 
For every observation at time $t$, we construct a 50-step target action chunk. 

\textbf{Dataset-specific normalization.} As embodiments differ in physical ranges and control units, we compute statistics independently for each dataset after applying the same dimension mapping, 50-step chunk construction, and relative-action conversion. For every valid state and action dimension, we estimate the 1st and 99th percentiles and apply
\begin{equation}
\tilde{x}=2\frac{x-q_{0.01}}{q_{0.99}-q_{0.01}+10^{-6}}-1,
\end{equation}
During training, these dataset-specific statistics place heterogeneous state and action values on a common numerical scale, enabling stable joint optimization across embodiments.

\section{Exploring Ego-Robot Co-Training Paradigms}

Rather than committing to a single recipe for leveraging egocentric data, we investigate three complementary co-training paradigms. 
First, \textbf{direct utilization} (Section~\ref{sec:direct-utilization}) incorporates ego data into joint training while separate action heads isolate domain-specific control outputs. 
Second, \textbf{explicit alignment} (Section~\ref{sec:explicit-alignment}) uses a progressive curriculum based on aligned demonstrations to transfer egocentric knowledge to robot control. 
Third, \textbf{world modeling} (Section~\ref{sec:world-modeling}) exploits future-state prediction, an objective naturally suited to the visual interaction structure of egocentric data. Evaluating these paradigms under a shared framework enables a systematic comparison of alternative mechanisms for ego-data pre-training.

\subsection{Main Architecture}

\begin{figure}[!t]
\centering
\includegraphics[width=0.95\linewidth]{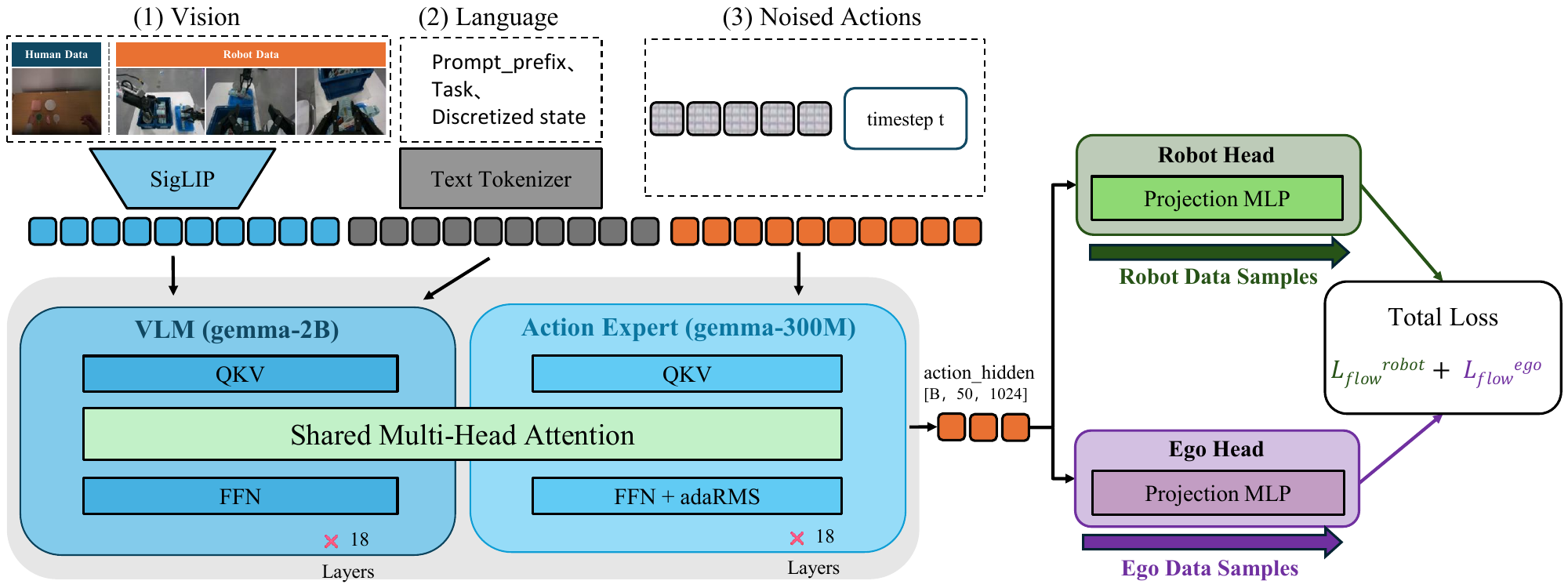}
\caption{Overview of joint ego--robot co-training with domain-specific action heads (Atom-DH). See Section~\ref{sec:direct-utilization} for details.}
\label{fig:atom-dh}
\end{figure}

AtomEgo investigates ego--robot co-training across both vision--language--action (VLA; Sections~\ref{sec:direct-utilization}--\ref{sec:explicit-alignment}) and world--action model (WAM; Section~\ref{sec:world-modeling}) architectures.

\textbf{Model instantiation.} The two VLA paradigms adopt the $\pi_{0.5}$ policy~\citep{pi05} architecture and are initialized from the base PaliGemma VLM weights rather than a pretrained $\pi_{0.5}$ policy checkpoint. 
The WAM paradigm instead retains the Fast-WAM architecture, as detailed in Section~\ref{sec:world-modeling}. Although their backbones and learning objectives differ, all paradigms use the same processed ego--robot corpus and standardized control interface.

\textbf{Model inputs and outputs.} Let $\mathcal{O}_t$ denote the current visual observation, organized into one base-view and two wrist-view slots; let $\ell$ denote the task instruction, $\mu$ the action-representation metadata, and $s_t$ the unified 80-dimensional robot state. At the policy interface, the two model families are expressed as
\begingroup
\setlength{\abovedisplayskip}{6pt}
\setlength{\belowdisplayskip}{6pt}
\setlength{\abovedisplayshortskip}{6pt}
\setlength{\belowdisplayshortskip}{6pt}
\begin{equation}
\hat{A}_{\mathrm{VLA}}=f_{\mathrm{VLA}}(\mathcal{O}_t,\ell,\mu,s_t),
\qquad
\hat{A}_{\mathrm{WAM}}=f_{\mathrm{WAM}}(\mathcal{O}_t,\ell),
\end{equation}
\endgroup
Both outputs are normalized action chunks in $\mathbb{R}^{B\times50\times80}$, representing 50 future control steps in the action space.

\subsection{Atom-DH: Joint Co-Training with Domain-Specific Action Heads}
\label{sec:direct-utilization}

Egocentric and robot trajectories provide complementary interaction experience, but their control targets follow distinct embodiment- and domain-specific distributions. 
A fully shared policy must therefore fit heterogeneous action mappings through the same output projection, which can introduce conflicting supervision at the control interface. 
We address this issue with a minimal architectural modification: the perception, language, and action representations are shared across domains, while the final action projection is separated into ego-specific and robot-specific heads.

Let $h_i\in\mathbb{R}^{50\times D}$ denote the action representation produced by the shared backbone for sample $i$, and let $d_i\in\{\mathrm{ego},\mathrm{robot}\}$ denote its domain. Two independently parameterized projections decode the shared representation,
\begingroup
\setlength{\abovedisplayskip}{6pt}
\setlength{\belowdisplayskip}{6pt}
\setlength{\abovedisplayshortskip}{6pt}
\setlength{\belowdisplayshortskip}{6pt}
\begin{equation}
\hat{v}_i =
\begin{cases}
W_{\mathrm{ego}}h_i, & d_i=\mathrm{ego},\\
W_{\mathrm{robot}}h_i, & d_i=\mathrm{robot}.
\end{cases}
\end{equation}
\endgroup

Each sample supervises only its corresponding head, while gradients from both domains continue to update the shared backbone. 
The ego head is used only to provide auxiliary supervision during co-training; downstream robot control is decoded exclusively through the robot head. 

We additionally evaluate an optimal-transport (OT) regularizer on the aligned human--robot subset. It derives temporal correspondences from end-effector trajectories and encourages matched action latents to remain close, while a lightweight embodiment prompt distinguishes human and robot inputs. Implementation details and experimental results for this variant are provided in Appendix~\ref{app:ot-variant}.

Overall, the dual-head model provides a minimal baseline for exploiting egocentric data through shared representation learning while retaining domain-specific action decoding. It also serves as a controlled reference for the more structured transfer paradigms introduced below.

\subsection{Atom-CL: Progressive Ego-to-Robot Transfer via Embodiment Alignment}

\begin{figure}[!t]
\centering
\includegraphics[width=0.95\linewidth]{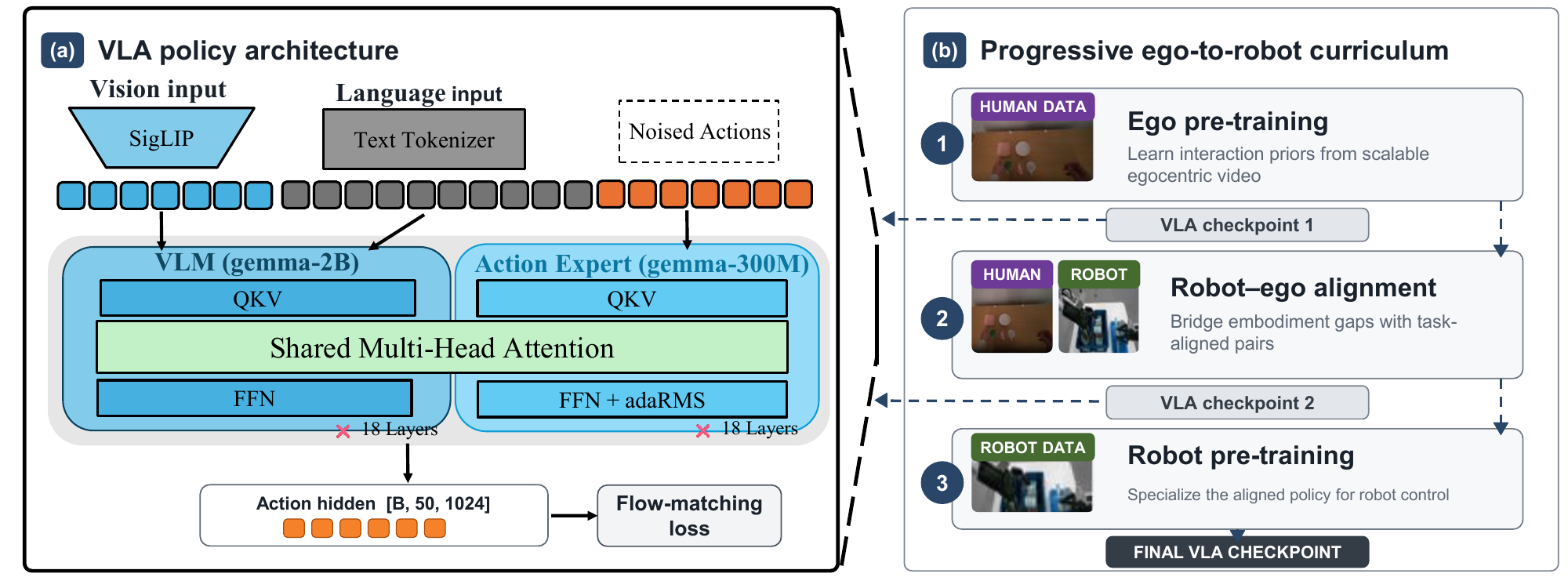}
\caption{Overview of progressive ego-to-robot transfer through curriculum-based embodiment alignment (Atom-CL). See Section~\ref{sec:explicit-alignment} for details.}
\label{fig:atom-cl}
\end{figure}

\begin{table*}[!t]
\centering
\caption{Three-stage pre-training curriculum for progressive ego-to-robot transfer.}
\label{tab:progressive-transfer}
\scriptsize
\setlength{\tabcolsep}{3pt}
\begin{tabular}{@{}lp{5.0cm}p{3.5cm}rp{3.2cm}@{}}
\toprule
Stage & Objective and training data & Initialization & Updates & Trainable components \\
\midrule
1 & Egocentric prior learning on Ego data & Base PaliGemma VLM & 100,000 & Full model \\
2 & Embodiment alignment on task-aligned human and robot data & Stage~1 checkpoint & 50,000 & Vision encoder and action expert \\
3 & Robot pre-training on robot datasets & Stage~2 checkpoint & 97,728 & Full model \\
\bottomrule
\end{tabular}
\end{table*}

\label{sec:explicit-alignment}

Directly mixing egocentric and robot supervision requires the model to bridge a substantial embodiment gap within a single training stage.
Prior work has shown that introducing an intermediate alignment stage can improve the utilization of egocentric data~\citep{egoscale2026}; we further investigate this strategy at pre-training scale.
We formulate ego-to-robot transfer as a curriculum with three successive data distributions: 
The model first learns general interaction priors from large-scale egocentric trajectories, then adapts on a smaller set of task-aligned human and robot demonstrations, and finally trains on robot pre-training data. 
The intermediate alignment stage provides a gradual transition from human interaction patterns to robot behavior, enabling the three-stage curriculum to fully leverage the benefits of egocentric data.

\paragraph{Aligned human--robot supervision.}
The alignment set contains human and robot demonstrations that share task semantics and interaction objectives, while differing in morphology, appearance, and native control space. 
Within this stage, we convert both domains to a common end-effector representation. 
A tracked human hand pose is treated as a canonical end effector, while robot joint trajectories are converted to Cartesian end-effector poses using embodiment-specific forward kinematics when Cartesian poses are not directly available. 

Both are expressed in their fixed egocentric camera frame and then transformed into motion relative to the current end-effector frame. Let ${}^{C}\!T_{E_t}$ denote the end-effector pose at time $t$ in camera frame $C$. For prediction horizon $\tau_h$, the aligned target is the relative transform from the current pose to the future pose,
\begingroup
\setlength{\abovedisplayskip}{6pt}
\setlength{\belowdisplayskip}{6pt}
\setlength{\abovedisplayshortskip}{6pt}
\setlength{\belowdisplayshortskip}{6pt}
\begin{equation}
\Delta T_{t,h}
=
\left({}^{C}\!T_{E_t}\right)^{-1}{}^{C}\!T_{E_{t+\tau_h}}.
\end{equation}
\endgroup
This construction aligns ego and robot trajectories at the representation level, allowing the task-matched dataset to provide an explicit transition from egocentric pre-training to robot learning.
This common representation is used specifically for the alignment stage; the subsequent robot-only stage retains each robot dataset's original joint-space control targets. Consequently, only the small alignment set requires additional conversion, avoiding costly reconstruction of the full pre-training corpus.

\paragraph{Progressive training.}
The pre-training curriculum consists of three sequential stages. 
Egocentric pre-training first acquires broad interaction priors, alignment fine-tuning then bridges human and robot motion under the shared end-effector representation, and robot pre-training finally adapts the aligned model to heterogeneous executable control. 
Each stage initializes from a checkpoint of the preceding stage. 
Table~\ref{tab:progressive-transfer} summarizes this schedule.

Overall, this design offers two advantages: progressive transfer of egocentric priors and lower processing cost, since alignment-specific post-processing is required only for the second-stage alignment dataset, not the full corpus.

\subsection{Atom-WAM: Joint World-Action Modeling for Cross-Embodiment Transfer}

\begin{figure}[t]
\centering
\includegraphics[width=0.95\linewidth]{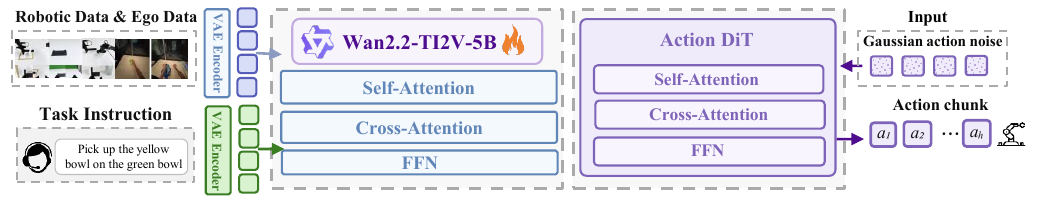}
\caption{Overview of joint world--action modeling for ego--robot co-training (Atom-WAM). See Section~\ref{sec:world-modeling} for details.}
\label{fig:atom-wam}
\end{figure}

\label{sec:world-modeling}

The preceding paradigms address the embodiment gap at the action interface, either by separating domain-specific outputs or by explicitly aligning human and robot motions. 
Our third paradigm instead exploits a signal that is naturally shared across embodiments: the visual evolution of an interaction. Although human hands and robot manipulators have different control spaces, both induce temporally structured changes in objects and scenes. 
We therefore instantiate this route by fine-tuning Fast-WAM~\citep{fastwam} on a mixture of egocentric and robot trajectories, using future-video modeling as a shared channel for cross-embodiment transfer.

\paragraph{Fast-WAM initialization.}
Unlike the two VLA-based routes above, this paradigm initializes from Fast-WAM architecture. Fast-WAM couples a video Diffusion Transformer (DiT)~\citep{DiT} with an Action DiT under a shared-attention Mixture-of-Transformers formulation. The Video DiT models future visual latents, while the Action DiT predicts a continuous action chunk from the current observation and language instruction. During training, clean observation tokens provide a common visual anchor for both branches. A structured attention mask allows both branches to access this anchor while preventing action tokens from attending to noisy future-video latent tokens, thereby avoiding future-information leakage.

\paragraph{Joint ego--robot fine-tuning.}
We train the model on the curated robot and egocentric data described in Section~3. We construct the fine-tuning distribution as
\begin{equation}
\mathcal{D}_{\mathrm{mix}}
=\rho\mathcal{D}_{\mathrm{ego}}+(1-\rho)\mathcal{D}_{\mathrm{robot}},
\end{equation}
where $\rho$ is the effective ego-data sampling probability. Future frames from both domains supervise the video flow-matching objective. Following Fast-WAM, we optimize
\begin{equation}
\mathcal{L}_{\mathrm{joint}}
=\mathbb{E}_{\mathcal{D}_{\mathrm{mix}}}
\left[\mathcal{L}_{\mathrm{action}}
+\lambda_{\mathrm{video}}\mathcal{L}_{\mathrm{video}}\right],
\end{equation}
where $\lambda_{\mathrm{video}}$ balances action learning and video co-training. In this way, both domains update the world representation through future-video prediction, while action supervision is restricted to the control dimensions available for each sample. The shared visual-dynamics objective therefore provides the primary transfer channel between human and robot embodiments without requiring their native action spaces to be identical.

At deployment, we follow the Fast-WAM inference procedure and omit future-video tokens, denoising, and decoding. The Video DiT processes only the current visual context to produce the latent world representation, which then generates the robot action chunk.

\subsection{Training pipeline}

All three paradigms follow a common pre-training--post-training--evaluation pipeline. Our study focuses on pre-training, where the methods differ in how they incorporate egocentric and robot data. Each pre-training resulting checkpoint is subsequently post-trained on the same in-house multi-task robot dataset using an identical recipe, isolating the effect of the pre-training strategy. All pre-training runs use the same training steps, batch size, and other hyperparameters; each uses $3\times8$ NVIDIA B200 GPUs and takes approximately four days.

\setcounter{section}{4}
\section{Experiments}

\begin{figure*}[!t]
\centering
\includegraphics[width=\textwidth]{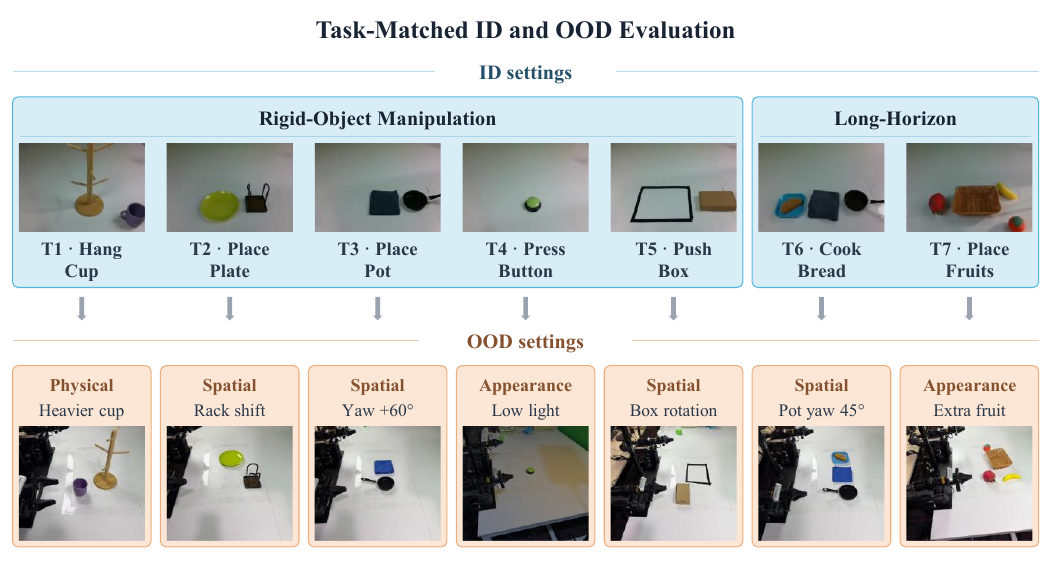}
\caption{Task-matched ID and OOD evaluation settings for the seven real-robot tasks. Each OOD setting preserves the task objective while changing one primary spatial, appearance, or physical factor.}
\label{fig:real-robot-id-ood-settings}
\end{figure*}

\label{sec:evaluation}
Having introduced three paradigms for incorporating egocentric data during pre-training, we now evaluate their effectiveness and derive practical insights. 
Real-robot task performance provides the most direct evaluation; in Section~\ref{sec:real-robot-evaluation}, we evaluate a suite of manipulation tasks under both in-distribution and out-of-distribution conditions. 
Beyond real-robot evaluation, Section~\ref{sec:cross-embodiment-representation} examines the latent representations of egocentric and robot data, providing an offline measure of cross-embodiment representation learning.

\subsection{Real-Robot Evaluation}
\label{sec:real-robot-evaluation}

\paragraph{Experimental setup.}
We conduct real-world experiments on a dual-arm platform; hardware details are provided in Appendix~\ref{app:hardware-platform}.

We evaluate seven manipulation tasks: \textit{Cook Bread},\textit{Hang Cup}, \textit{Place Fruits}, \textit{Place Plate},
\textit{Place Pot}, \textit{Press Button}, and \textit{Push Box}. 
For each method, we use only its final pre-training checkpoint, obtained after approximately 100,000 updates, for identical joint post-training on all seven tasks; the resulting post-training checkpoint is evaluated under the multi-task setting.

For each checkpoint, we evaluate all seven tasks, conducting ten trials per task under the nominal ID condition and ten under the corresponding OOD condition, yielding 140 trials in total.
Each OOD variant preserves the task objective while changing one primary spatial, appearance, or physical factor. 
A rollout is counted as successful only when it reaches the complete predefined task end state.
The matched nominal and shifted task settings are summarized in Figure~\ref{fig:real-robot-id-ood-settings}.

\paragraph{Baselines.}
Each baseline is pre-trained only on robot demonstrations using the standard objective, without ego-specific mechanisms, then post-trained on the same seven-task robot dataset.
In contrast, the Atom variants incorporate both robot and ego data during pre-training through their respective co-training strategies, while following the identical post-training procedure. 
All methods use the same training steps, learning rate, batch size, and other hyperparameters.
Because the VLA and WAM routes use different model architectures, we construct a separate robot-only baseline for each family. The VLA baseline is compared with domain-specific dual-head co-training (\textbf{Atom-DH}) and progressive embodiment alignment (\textbf{Atom-CL}), while the WAM baseline is compared with joint world--action co-training (\textbf{Atom-WAM}).

\begin{table*}[t]
\centering
\caption{Per-task real-robot results under ID and OOD conditions.}
\label{tab:per-task-robot-results}
\scriptsize
\setlength{\tabcolsep}{2.5pt}
\renewcommand{\arraystretch}{1.08}
\scalebox{1.125}[1]{%
\resizebox{0.8\textwidth}{!}{%
\begin{tabular}{@{}l*{10}{c}@{}}
\toprule
& \multicolumn{6}{c}{VLA} & \multicolumn{4}{c}{WAM} \\
\cmidrule(lr){2-7}\cmidrule(lr){8-11}
& \multicolumn{2}{c}{Baseline} & \multicolumn{2}{c}{Atom-DH} & \multicolumn{2}{c}{Atom-CL}
& \multicolumn{2}{c}{Baseline} & \multicolumn{2}{c}{Atom-WAM} \\
\cmidrule(lr){2-3}\cmidrule(lr){4-5}\cmidrule(lr){6-7}\cmidrule(lr){8-9}\cmidrule(lr){10-11}
Task & ID & OOD & ID & OOD & ID & OOD & ID & OOD & ID & OOD \\
\midrule
Overall (7 tasks) & 30/70 & 16/70 & 32/70 & 29/70 & 45/70 & 30/70 & 21/70 & 19/70 & 15/70 & 14/70 \\
Success rate & 42.86\% & 22.86\% & 45.71\% & 41.43\% & 64.29\% & 42.86\% & 30.00\% & 27.14\% & 21.43\% & 20.00\% \\
\midrule
Cook Bread & \textbf{10/10} & 0/10 & 1/10 & 0/10 & 2/10 & 0/10 & 2/10 & \textbf{1/10} & 0/10 & 0/10 \\
Hang Cup & 1/10 & 1/10 & 0/10 & 0/10 & \textbf{10/10} & \textbf{10/10} & 0/10 & 0/10 & 0/10 & 0/10 \\
Place Fruits & 9/10 & 8/10 & \textbf{10/10} & 9/10 & \textbf{10/10} & \textbf{10/10} & 9/10 & \textbf{10/10} & \textbf{10/10} & \textbf{10/10} \\
Place Plate & 0/10 & 0/10 & 0/10 & 0/10 & \textbf{1/10} & 0/10 & \textbf{1/10} & 0/10 & 0/10 & 0/10 \\
Place Pot & \textbf{4/10} & \textbf{2/10} & 1/10 & 0/10 & 2/10 & 0/10 & 0/10 & 0/10 & 0/10 & 0/10 \\
Press Button & 1/10 & 1/10 & \textbf{10/10} & \textbf{10/10} & \textbf{10/10} & \textbf{10/10} & 9/10 & 8/10 & 5/10 & 4/10 \\
Push Box & 5/10 & 4/10 & \textbf{10/10} & \textbf{10/10} & \textbf{10/10} & 0/10 & 0/10 & 0/10 & 0/10 & 0/10 \\
\bottomrule
\end{tabular}
}
}
\vspace{8pt}
\begin{minipage}{1.0\textwidth}
\centering
\includegraphics[width=\linewidth]{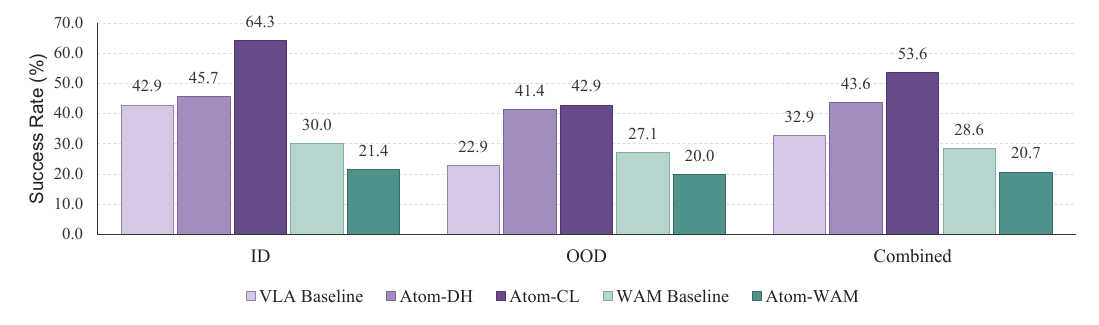}
\captionof{figure}{Real-robot success rates of the VLA and WAM checkpoints under ID, OOD, and combined evaluation settings.}
\label{fig:robot-success-rates}
\end{minipage}
\vspace{-10pt}
\end{table*}

\paragraph{Results.}
Table~\ref{tab:per-task-robot-results} reports the aggregate and per-task real-robot results, which are summarized in Figure~\ref{fig:robot-success-rates}. 
Within the VLA family, Atom-CL achieves the highest ID and combined success rates of 64.29\% and 53.57\%, improving over the VLA baseline by 21.43 and 20.71 percentage points, respectively. Atom-DH reaches 43.57\% combined success. Both methods improve OOD success over the VLA baseline's 22.86\%, reaching 41.43\% for Atom-DH and 42.86\% for Atom-CL.
In contrast, Atom-WAM surprisingly underperforms its WAM baseline across both conditions: ID success decreases from 30.00\% to 21.43\%, OOD success decreases from 27.14\% to 20.00\%, and combined success decreases from 28.57\% to 20.71\%.
This counterintuitive result suggests that visual prediction alone may capture domain-specific dynamics without sufficiently grounding them in executable robot actions.
Together, the results indicate that egocentric data are not inherently beneficial and that their value depends strongly on the integration strategy, with progressive embodiment alignment providing the most consistent gains under this evaluation.
While this discussion focuses on the direct experimental results, we will provide a higher-level synthesis and overall takeaways in Section~\ref{sec:result-summary}.


\begin{table*}[t]
\centering
\caption{Language-conditioned cross-embodiment representation results on seven tasks. Higher raw matched-task similarity indicates stronger consistency, while lower normalized different-task similarity indicates better task separation.}
\label{tab:seven-task-semantic-results}
\small
\begin{tabular}{lcc}
\toprule
Checkpoint & Raw matched mean $\uparrow$ & Normalized different-task mean $\downarrow$ \\
\midrule
Atom-DH & $\mathbf{0.999625}$ & $0.592719$ \\
Atom-CL & $0.994481$ & $\mathbf{0.387240}$ \\
Atom-WAM & $0.997357$  & $0.480727$ \\
\bottomrule
\end{tabular}
\vspace{8pt}
\begin{minipage}{\textwidth}
\centering
\includegraphics[width=\linewidth]{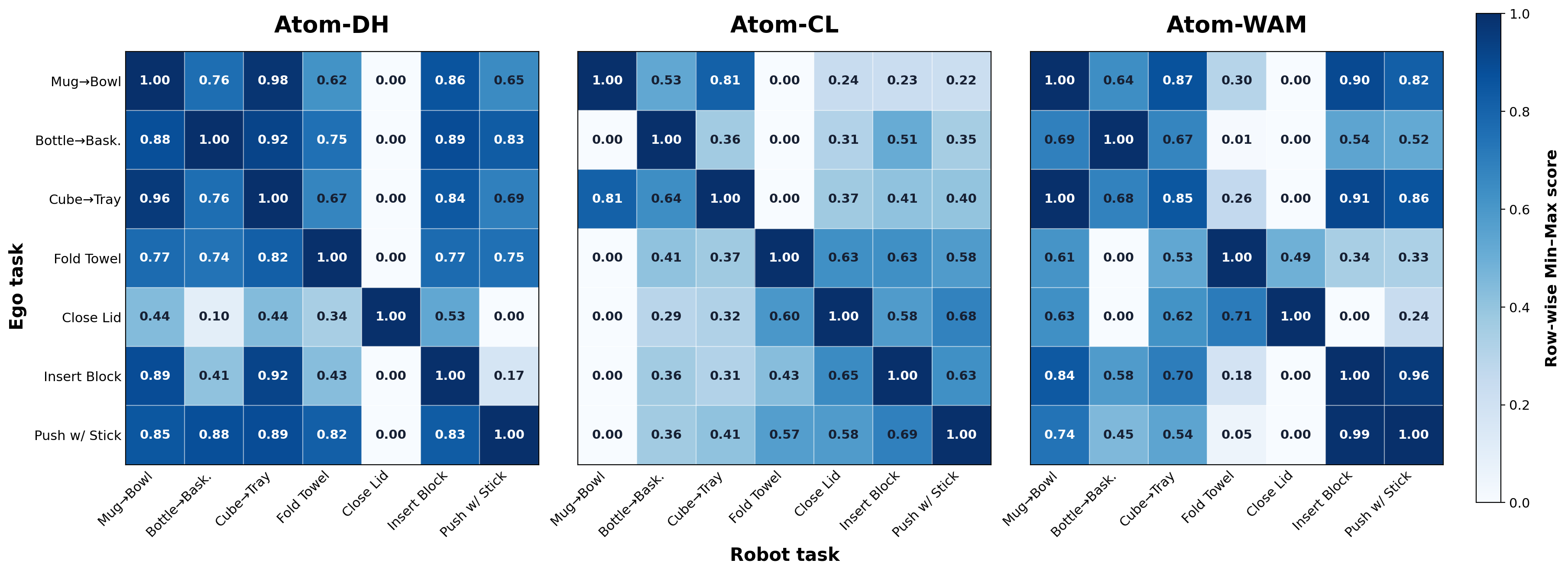}
\captionof{figure}{Row-normalized language-conditioned Ego-to-Robot task similarity for Atom-DH (top left), Atom-CL (top right), and Atom-WAM (bottom). Rows denote Ego tasks and columns denote Robot tasks.}
\label{fig:seven-task-heatmaps}
\end{minipage}
\end{table*}

\subsection{Language-Conditioned Cross-Embodiment Representation Evaluation}
\label{sec:cross-embodiment-representation}

To further examine cross-embodiment representation learning, we evaluate whether ego and robot videos of the same task are encoded similarly while remaining distinguishable from other tasks. We use seven task-matched sets from our self-collected alignment dataset, which contains egocentric and real-robot demonstrations of the same tasks.


\paragraph{Protocol and metrics.}
We evaluate three post-trained models---Atom-DH, Atom-CL, and Atom-WAM---on 62 ego and 62 robot videos. For each video, we uniformly sample 32 front-view frames, pair them with the task instruction, and aggregate the resulting language-conditioned frame features into a clip representation $z$ using temporal-change weighting.
For ego task $p$ and robot task $q$, we compute
\begin{equation}
S_{pq}
=
\frac{1}{|\mathcal{E}_p||\mathcal{R}_q|}
\sum_{i\in\mathcal{E}_p}\sum_{j\in\mathcal{R}_q}
\operatorname{cos}(z_i,z_j),
\end{equation}
where $\mathcal{E}_p$ and $\mathcal{R}_q$ denote the corresponding ego and robot video sets. The resulting $7\times7$ matrix is visualized in Figure~\ref{fig:seven-task-heatmaps} after row-wise Min--Max normalization.
Table~\ref{tab:seven-task-semantic-results} reports the mean raw diagonal similarity for matched tasks and the mean normalized off-diagonal similarity for different tasks. Together, these metrics measure cross-embodiment task consistency and task discrimination.

\paragraph{Results.}
Atom-DH achieves the highest matched-task similarity but also the highest normalized different-task similarity, indicating strong cross-embodiment consistency alongside relatively weak task separation.
Atom-CL exhibits slightly lower matched-task similarity but substantially better task separation, while Atom-WAM lies between the two VLA methods on both metrics.
This intermediate position suggests that joint visual prediction captures some shared cross-embodiment structure but does not produce the strongest task selectivity.
More broadly, the representation ordering does not directly match the real-robot performance ordering.
Useful cross-embodiment alignment must therefore preserve action-relevant task structure rather than simply bringing ego and robot representations closer.

\subsection{Result Summary and Takeaways}
\label{sec:result-summary}
We now return to our initial question: How can we use egocentric data effectively during pre-training? Our experiments yield the following takeaways.

\paragraph{Takeaway 1: Effective alignment requires more than direct data mixing.}
Simply combining egocentric and robot data during pre-training does not reliably improve robot control. Despite providing greater scale and interaction diversity, the substantial embodiment and action-space gaps can introduce conflicting supervision. Consistent with this observation, Atom-DH, which relies only on domain-specific action-head separation, improves ID success only marginally, from 42.86\% to 45.71\%. Thus, effective transfer requires structured cross-embodiment alignment rather than direct data mixing.

\paragraph{Takeaway 2: Ego data particularly benefit OOD generalization.}
Both ego-pretrained VLA variants improve OOD success over the VLA baseline's 22.86\%, reaching 41.43\% for Atom-DH and 42.86\% for Atom-CL. These gains suggest that diverse human interactions can provide transferable priors that are particularly valuable under distribution shift, enabling generalization beyond the robot training distribution.


\paragraph{Takeaway 3: WAM-based transfer requires stronger alignment mechanisms.}
Direct ego--robot co-training with Fast-WAM~\cite{fastwam} reduces combined success from 28.57\% to 20.71\%. Although both domains share the same future-video supervision objective, this common visual objective alone does not bridge the embodiment gap. Effective ego utilization in WAMs therefore requires finer-grained cross-embodiment alignment rather than naive data mixing, or a stronger, carefully designed architecture rather than simply reusing Fast-WAM.

Overall, our findings suggest a simple principle: \emph{Data Scale $\times$ Alignment Quality $\rightarrow$ Capability Gain}; the value of egocentric data depends not only on its scale, but also on how effectively it is aligned and utilized.

\section{Limitation and future work}
While this study provides a controlled comparison of representative ego--robot co-training paradigms, several directions remain open. First, due to resource constraints, the scale of our pre-training corpus remains limited. Second, our scope prioritizes a systematic evaluation of alternative paradigms rather than specialized algorithmic development for any single route. 
In the future, we plan to further investigate the most effective approach and scale its training to substantially larger datasets toward a more capable embodied foundation model.

\section{Conclusion}
We presented AtomEgo, a unified framework for studying scalable ego--robot co-training for embodied foundation models. 
Using a curated corpus of robot and egocentric interaction data, we compared three complementary paradigms. 
Our results show that egocentric data can improve policy generalization, but their effectiveness depends critically on the cross-embodiment alignment strategy. 
These findings provide a practical foundation for scaling embodied models beyond robot-only supervision.

\newpage

\section*{Acknowledgements}
We gratefully acknowledge the computational support provided by the Lionrock Artificial Intelligence Laboratory at the China Merchants Research Institute of Advanced Technology.

\section*{Author Contributions}
\noindent\textbf{Data collection and processing:} Di Wu.\par
\noindent\textbf{Algorithm and training:} Junhe Sheng, Zhongxing Wei, Xiaoquan Sun, Junyang Zheng, Di Wu, Songxin Zhang and Zejian Xie.\par
\noindent\textbf{Evaluation:} Dongchen Zheng.\par
\noindent\textbf{Writing:} Di Wu and Dongchen Zheng.\par
\noindent\textbf{Advisor:} Jiayu Chen, Jiaxing Zhang, Zhuoyang Song.

\bibliographystyle{plainnat}
\bibliography{references}

\clearpage
\appendix
\section*{Appendix Contents}
\begingroup
\small
\setlength{\parskip}{2pt}
\noindent\hyperref[app:evaluation-details]{\textbf{\ref*{app:evaluation-details}\quad\nameref*{app:evaluation-details}}}\hfill\pageref{app:evaluation-details}\par
\noindent\hspace*{1.5em}\hyperref[app:hardware-platform]{\ref*{app:hardware-platform}\quad\nameref*{app:hardware-platform}}\hfill\pageref{app:hardware-platform}\par
\noindent\hspace*{1.5em}\hyperref[app:cross-embodiment-instructions]{\ref*{app:cross-embodiment-instructions}\quad\nameref*{app:cross-embodiment-instructions}}\hfill\pageref{app:cross-embodiment-instructions}\par
\noindent\hspace*{1.5em}\hyperref[app:task-suite]{\ref*{app:task-suite}\quad\nameref*{app:task-suite}}\hfill\pageref{app:task-suite}\par
\noindent\hyperref[app:training-hyperparameters]{\textbf{\ref*{app:training-hyperparameters}\quad\nameref*{app:training-hyperparameters}}}\hfill\pageref{app:training-hyperparameters}\par
\noindent\hyperref[app:ot-variant]{\textbf{\ref*{app:ot-variant}\quad\nameref*{app:ot-variant}}}\hfill\pageref{app:ot-variant}\par
\endgroup

\section{Evaluation Details}
\label{app:evaluation-details}

\subsection{Hardware Platform}
\label{app:hardware-platform}

\begin{figure}[htbp]
\centering
\includegraphics[width=0.82\columnwidth]{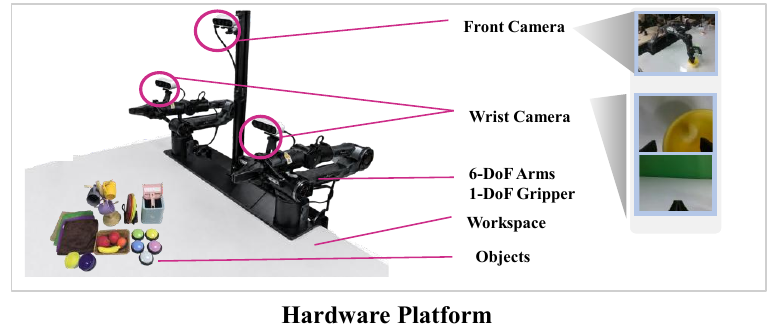}
\caption{Hardware platform used for real-robot evaluation.}
\label{fig:hardware-platform}
\end{figure}

All real-robot experiments use a dual-arm platform comprising two AgileX Piper 6-DoF arms. The system is equipped with three Intel RealSenseD435i RGB cameras, including two wrist cameras, and one headcamera. Figure~\ref{fig:hardware-platform} shows the hardware setup used throughout the seven-task evaluation.

\subsection{Cross-Embodiment Evaluation Instructions}
\label{app:cross-embodiment-instructions}

The cross-embodiment representation evaluation uses the following seven fixed
task instructions in the same order as the evaluation matrix. The Ego and
Robot videos associated with the same task receive exactly the same
instruction.

\begin{enumerate}
    \item \texttt{pink\_mug\_to\_green\_bowl}: ``Pick up the pink mug and place it in the green bowl.''
    \item \texttt{water\_bottle\_to\_basket}: ``Pick up the water bottle and place it in the basket.''
    \item \texttt{red\_cube\_into\_blue\_tray}: ``Pick up the red cube and place it in the blue tray.''
    \item \texttt{fold\_towel}: ``Use both hands to grasp the two designated corners along the lower edge of the towel and fold the towel once from bottom to top.''
    \item \texttt{close\_box\_lid}: ``Use both hands to grasp the two sides of the box lid and close the lid.''
    \item \texttt{insert\_block}: ``Pick up the wooden block and insert it into the hole of the white foam block.''
    \item \texttt{push\_cube\_with\_stick}: ``Pick up the wooden stick, then use it to push the red cube completely into the white target area.''
\end{enumerate}

\subsection{Real-Robot Task Definitions}
\label{app:task-suite}

The ID configurations follow the nominal setups used in our seven-task real-robot evaluation. Each paired OOD condition preserves the task objective while perturbing one primary factor, allowing the resulting performance change to be associated with a specific type of distribution shift. Task initializations, perturbation types and execution timing, maximum rollout durations, and complete-task success criteria are fixed before evaluation and shared across checkpoints. Table~\ref{tab:task-suite} lists the seven evaluated task configurations and success criteria.

\begin{table}[H]
\centering
\caption{Seven-task real-robot evaluation protocol. Each OOD condition changes one primary factor while holding other task-relevant variables fixed whenever applicable. Symbolic quantities denote protocol parameters fixed before execution, including illumination, mass, displacement, and stability duration.}
\label{tab:task-suite}
\scriptsize
\setlength{\tabcolsep}{2pt}
\renewcommand{\arraystretch}{1.12}
\begin{tabular}{@{}>{\raggedright\arraybackslash}p{1.35cm}
                    >{\raggedright\arraybackslash}p{1.60cm}
                    >{\raggedright\arraybackslash}p{1.20cm}
                    >{\raggedright\arraybackslash}p{1.80cm}
                    >{\raggedright\arraybackslash}p{2.80cm}
                    >{\raggedright\arraybackslash}p{4.20cm}@{}}
\toprule
Task & ID setup & OOD type & OOD perturbation & OOD specification and controls & Success criterion \\
\midrule
Cook Bread & Standard pot and bread poses & Spatial & Rotate pot & Pot yaw $=45^\circ$ CCW; pot center and other object poses unchanged & Bread is placed in the pot, and the pot is placed on the cloth \\
\addlinespace
Hang Cup & Standard cup mass & Physical & Increase cup mass & Mass $m_{\rm ID}\!\rightarrow m_{\rm OOD}$; cup appearance and initial pose unchanged & Handle is inserted and the cup hangs stably for the predefined duration \\
\addlinespace
Place Fruits & Standard target set & Appearance & Add visually similar non-target fruit as a distractor & Distractor identity and pose fixed; target identities and initial poses unchanged & All three target fruits are placed in the basket while the distractor remains outside \\
\addlinespace
Place Plate & Standard rack pose & Spatial & Translate rack & Rack displacement $\Delta\mathbf{x}_{\rm rack}$ fixed; plate pose and rack orientation unchanged & Plate is aligned, inserted, released, and remains stable in the rack \\
\addlinespace
Place Pot & Standard pot pose & Spatial & Rotate pot & Pot yaw $=60^\circ$; pot center and cloth pose unchanged & Pot is grasped by the handle and placed on the cloth \\
\addlinespace
Press Button & Standard illumination & Appearance & Reduce illumination & Illuminance $L_{\rm ID}\!\rightarrow L_{\rm OOD}$; camera and button poses unchanged & Button activation is registered and the end effector reaches the predefined retreat region \\
\addlinespace
Push Box & Standard box orientation & Spatial & Rotate box & Box yaw $=90^\circ$; box center and box--target distance unchanged & Box reaches the predefined target region \\
\bottomrule
\end{tabular}
\end{table}

\begin{figure*}[p]
\centering
\includegraphics[width=\textwidth,height=0.82\textheight,keepaspectratio]{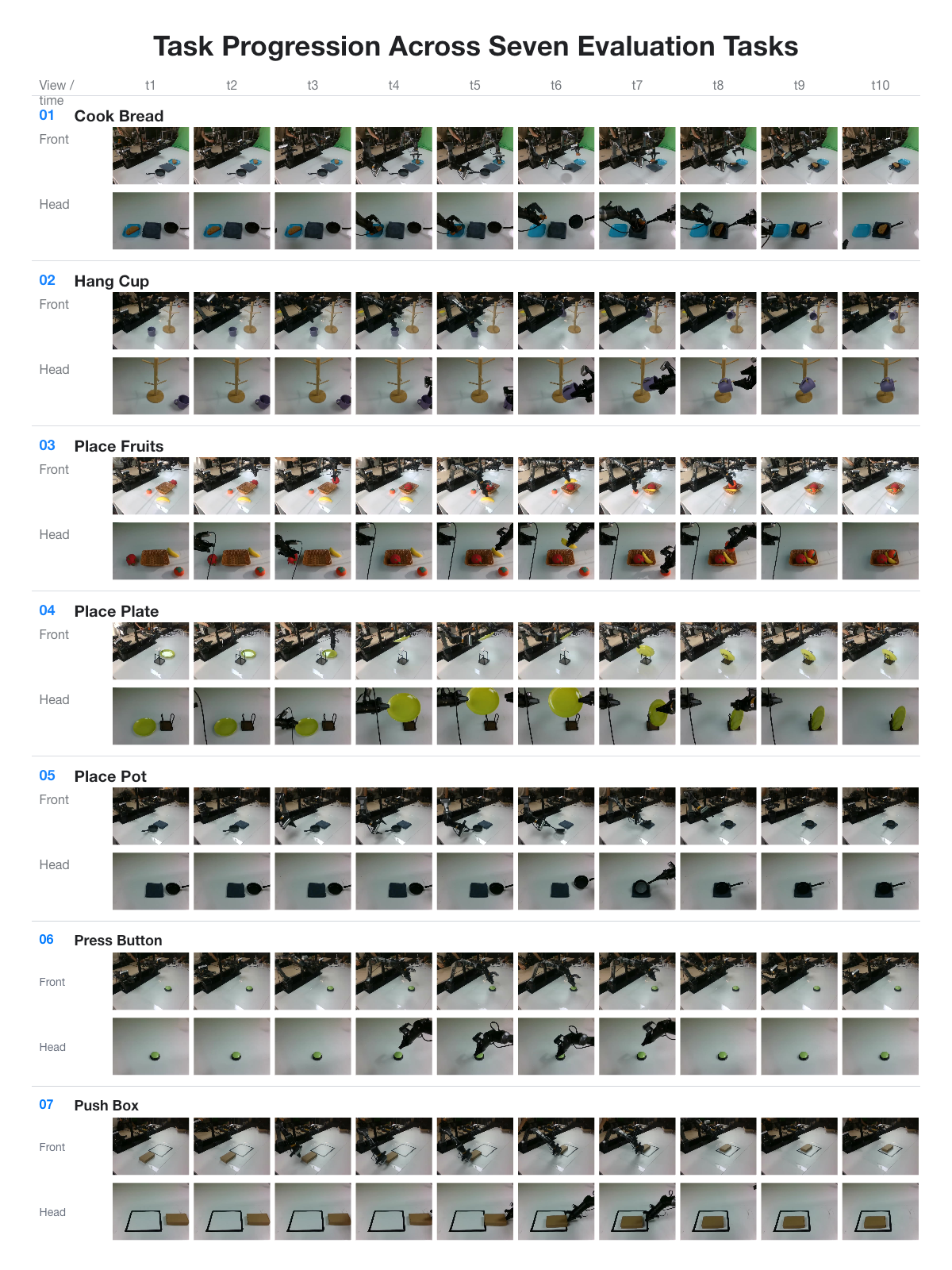}
\caption{Uniformly sampled task-completion progressions for the seven real-robot evaluation tasks. Each row pair shows front and head views at ten time points from a complete task execution.}
\label{fig:real-robot-task-progressions}
\end{figure*}

Figure~\ref{fig:real-robot-task-progressions} provides qualitative task-completion progressions for the same seven-task evaluation suite.

\section{Training Hyperparameters}
\label{app:training-hyperparameters}

Table~\ref{tab:training-hyperparameters} summarizes the key settings for robot pre-training and subsequent multi-task post-training. Pre-training uses 34 audited in-house and open-source robot datasets, while post-training adapts the model to two in-house Piper multi-task datasets. Under the reported hardware configuration, pre-training takes approximately four days, while post-training takes about 12 hours.

\begin{table}[H]
\centering
\caption{Key hyperparameters for robot pre-training and multi-task post-training.}
\label{tab:training-hyperparameters}
\footnotesize
\setlength{\tabcolsep}{4pt}
\renewcommand{\arraystretch}{1.08}
\begin{tabular}{@{}>{\raggedright\arraybackslash}p{3.6cm}
                    >{\raggedright\arraybackslash}p{4.7cm}
                    >{\raggedright\arraybackslash}p{4.7cm}@{}}
\toprule
Hyperparameter & Robot pre-training & Multi-task post-training \\
\midrule
Training data & 34 robot datasets; 150.1M source frames; EgoVerse excluded & Two in-house Piper multi-task datasets \\
Action representation & Unified80 & Unified80 \\
Hardware & $3\times8$ NVIDIA B200 GPUs & $1\times8$ NVIDIA B200 GPUs \\
FSDP devices & 4 & 4 \\
Global batch size (train / validation) & 1,536 / 96 & 512 / 96 \\
Training steps & 97,728 & 20,000 \\
Warmup / decay steps & 5,000 / 97,728 & 1,000 / 30,000 \\
Peak / decay learning rate & $1\times10^{-6}$ / $1\times10^{-7}$ & $2.5\times10^{-5}$ / $2.5\times10^{-6}$ \\
Evaluation / checkpoint interval & 1,000 / 10,000 & 1,000 / 5,000 \\
Action MSE & Disabled & Enabled \\
Initialization & Base PaliGemma VLM & Final robot pre-training checkpoint \\
\bottomrule
\end{tabular}
\end{table}

\section{Optimal-Transport Alignment Details}
\label{app:ot-variant}

The domain-specific heads in Section~\ref{sec:direct-utilization} isolate human and robot action decoding, but they do not explicitly constrain the shared action representation across domains. We therefore examine a training-only optimal-transport (OT) regularizer that brings human and robot latents closer when their end-effector trajectories describe similar interactions. 

\paragraph{Data pools and alignment supervision.}
Because data with meaningful human--robot correspondence constitute only a small fraction of the full corpus, unconstrained sampling would rarely place sufficient paired examples in the same batch. 
We therefore divide the training mixture into a general co-training pool and an alignment pool, and reserve a fixed batch quota for each.
The alignment pool consists exclusively of our in-house aligned data: wearable-camera human demonstrations and robot executions collected for the same task set. 
The general pool contains all remaining egocentric and robot datasets and preserves the scale and diversity of the complete corpus. 
Only samples from the alignment pool receive the additional OT supervision, while samples from the general pool retain the original flow-matching objective.

\paragraph{Trajectory-aware OT alignment.}
Human and robot demonstrations of the same interaction may progress at different speeds, making direct frame-by-frame latent matching unreliable. We first align their ground-truth Cartesian trajectories using Soft Dynamic Time Warping (Soft-DTW), a smooth sequence-alignment procedure that permits nonlinear temporal correspondence. This operation is applied only to end-effector positions and is detached from gradient computation. It produces a correspondence weight matrix $W$ that indicates which portions of the human and robot trajectories should be compared.

We construct separate bridges for single-arm and bimanual interactions. The former uses the right end-effector position, while the latter jointly uses the left and right end-effectors. For each bridge, valid human--robot pairs are collected from the global batch. We then use the Sinkhorn algorithm, an entropy-regularized solver for soft optimal transport, to match their action-expert latents with pairwise cost
\begin{equation}
C_{ij}
=
\frac{1}{2}
\overline{\left\lVert h_i-r_j\right\rVert_2^2}
W_{ij},
\end{equation}
where $h_i$ and $r_j$ denote the human and robot latent trajectories, and the overline averages over temporal and feature dimensions. The valid single-arm and bimanual transport costs are summed to obtain $\mathcal{L}_{\mathrm{OT}}$.

The final training objective is
\begin{equation}
\mathcal{L}
=
\mathcal{L}_{\mathrm{FM}}
+
\alpha\mathcal{L}_{\mathrm{OT}},
\end{equation}
where $\alpha$ controls the strength of latent alignment and flow matching remains the primary action objective. OT is used only during training; the output heads and inference procedure are unchanged, so the variant introduces no deployment-time computation.

\paragraph{Experiment result.}
We compare two Atom-DH variants that differ in whether the OT regularizer is enabled, using the same seven-task real-robot evaluation protocol used in the main paper. Each variant is evaluated with ten ID and ten OOD trials per task. Table~\ref{tab:ot-ablation} reports the resulting success rates.

\begin{table}[H]
\centering
\caption{Ablation of the training-only OT regularizer on the seven-task real-robot evaluation. Each ID or OOD entry aggregates 70 trials; Combined pools all 140 trials.}
\label{tab:ot-ablation}
\small
\setlength{\tabcolsep}{8pt}
\begin{tabular}{@{}lccc@{}}
\toprule
Variant & ID success & OOD success & Combined success \\
\midrule
Atom-DH (+OT) & 38.57\% (27/70) & 31.43\% (22/70) & 35.00\% (49/140) \\
Atom-DH (w/o OT) & \textbf{44.29\% (31/70)} & \textbf{42.86\% (30/70)} & \textbf{43.57\% (61/140)} \\
\bottomrule
\end{tabular}
\end{table}

Disabling OT improves ID, OOD, and combined success by 5.72, 11.43, and 8.57 percentage points, respectively. Under this protocol, explicit latent transport alignment does not improve downstream control and produces the largest degradation under OOD conditions. We therefore treat OT as an auxiliary training variant rather than a default component of Atom-DH.

\end{document}